\documentclass[sigconf,nonacm]{acmart}
\AtBeginDocument{%
  \providecommand\BibTeX{{%
    \normalfont B\kern-0.5em{\scshape i\kern-0.25em b}\kern-0.8em\TeX}}}

\setcopyright{none}
\copyrightyear{ }
\acmYear{ }
\acmDOI{}
\usepackage{graphicx}

\acmBooktitle{}
\acmPrice{}
\acmISBN{}

\begin{document}

\title{Modeling the geospatial evolution of COVID-19 using spatio-temporal convolutional sequence-to-sequence neural networks}


\author{Mário Cardoso}
\email{mario.j.c.cardoso@tecnico.ulisboa.pt}
\affiliation{%
  \institution{INESC-ID / Instituto Superior Técnico}
  \city{Lisbon}
  \country{Portugal}
}

\author{André Cavalheiro}
\email{andre.cavalheiro@tecnico.ulisboa.pt}
\affiliation{%
  \institution{INESC-ID / Instituto Superior Técnico}
  \city{Lisbon}
  \country{Portugal}
}

\author{Alexandre Borges}
\email{alexandre.borges@tecnico.ulisboa.pt}
\affiliation{%
  \institution{INESC-ID / Instituto Superior Técnico}
  \city{Lisbon}
  \country{Portugal}
}

\author{Ana F. Duarte}
\email{filipadamasoduarte@tecnico.ulisboa.pt}
\affiliation{%
  \institution{CERENA / Instituto Superior Técnico}
  \city{Lisbon}
  \country{Portugal}
}

\author{Amílcar Soares}
\email{asoares@tecnico.ulisboa.pt}
\affiliation{%
  \institution{CERENA / Instituto Superior Técnico}
  \city{Lisbon}
  \country{Portugal}
}

\author{Maria João Pereira}
\email{maria.pereira@tecnico.ulisboa.pt}
\affiliation{%
  \institution{CERENA / Instituto Superior Técnico}
  \city{Lisbon}
  \country{Portugal}
}

\author{Nuno Jardim Nunes}
\email{nunojnunes@tecnico.ulisboa.pt}
\affiliation{%
  \institution{ITI / Instituto Superior Técnico}
  \city{Lisbon}
  \country{Portugal}
}

\author{Leonardo Azevedo}
\email{leonardo.azevedo@tecnico.ulisboa.pt}
\affiliation{%
  \institution{CERENA / Instituto Superior Técnico}
  \city{Lisbon}
  \country{Portugal}
}

\author{Arlindo L. Oliveira}
\email{arlindo.oliveira@tecnico.ulisboa.pt}
\affiliation{%
  \institution{INESC-ID / Instituto Superior Técnico}
  \city{Lisbon}
  \country{Portugal}
}

\renewcommand{\shortauthors}{ }

\begin{abstract}
Europe was hit hard by the COVID-19 pandemic and Portugal was one of the most affected countries, having suffered three waves in the first twelve months. Approximately between Jan 19th and Feb 5th 2021 Portugal was the country in the world with the largest incidence rate, with 14-days incidence rates per 100,000 inhabitants in excess of 1000. Despite its importance, accurate prediction of the geospatial evolution of COVID-19 remains a challenge, since existing analytical methods fail to capture the complex dynamics that result from both the contagion within a region and the spreading of the infection from infected neighboring regions. 

We use a previously developed methodology and official municipality level data from the Portuguese Directorate-General for Health (DGS), relative to the first twelve months of the pandemic, to compute an estimate of the incidence rate in each location of mainland Portugal. The resulting sequence of incidence rate maps was then used as a gold standard to test the effectiveness of different approaches in the prediction of the spatial-temporal evolution of the incidence rate. Four different methods were tested: a simple cell level autoregressive moving average (ARMA) model, a cell level vector autoregressive (VAR) model, a municipality-by-municipality compartmental SIRD model followed by direct block sequential simulation and a convolutional sequence-to-sequence neural network model based on the STConvS2S architecture. We conclude that the convolutional sequence-to-sequence neural network is the best performing method, when predicting the medium-term future incidence rate, using the available information.

\end{abstract}



\maketitle

\section{Introduction}

Modeling the spatio-temporal dynamics of the evolution of the COVID-19 pandemic is important for a variety of reasons, which include health service planning, adoption of containment measures, and allocation of resources. In fact, knowing the incidence rate in a given area enables public health officials and individuals to adopt the best strategies to contain the disease and reduce the risk of infection.

Regrettably, it is difficult to obtain data that accurately reflects the incidence rate at any given time and place, since statistics are collected at an aggregate level at the country, province, municipality, or city level. Therefore, detailed maps of the incidence rate are usually not available, in part because of data collection limitations but also because, in general, it is not possible to determine exactly where people became infected with COVID-19.

However, since contagion depends on individual and social behaviour and only happens when people are in close proximity, incidence rate maps remain an excellent tool to plan and act in order to contain an epidemic that propagates mainly through person-to-person contact. 
In previous work % by some of the authors of this paper 
the authors have proposed an incidence rate spatial model for COVID-19 in Portugal, based on a geostatistical framework and using the confirmed positive tested cases reported by the Portuguese Directorate-General for Health (DGS acronym, in Portuguese) \cite{Azevedo20}. These incidence rate maps are a very good approximation to the real incidence rate since they are derived from official data and require only mild assumptions about the way epidemics spread through a territory. 

After a brief review of related work, in section \ref{sec:related}, we describe the method used to derive the incidence rate maps in section \ref{sec:dataset}.
To test the accuracy of the modeling algorithms in the incidence rate data, we implemented and evaluated four different methods, described in section \ref{sec:methods}. The results obtained with these models are reported in section \ref{sec:results}.  Section \ref{sec:conclusions} presents the conclusions gathered from this work and proposes a few directions for future work.

\section{Related work \label{sec:related}}

In this section, we provide a brief overview of related work on geospatial modeling of disease risks, with a focus on the methods that take into account the continuous nature of the territories.

\subsection{Geospatial modeling of disease risk}

Data concerning diseases is commonly available by administrative regions. For many infectious diseases, statistics on the number of cases by countries, municipalities or other administrative regions are widely available. Thus, the most common approaches to map and visualize the risk of disease is through the use of choropleth maps, where those pre-defined spatial regions are colored in accordance with recorded incidence rates. In the wake of the COVID-19 pandemic, many such analyses have been performed and made available, e.g,  \cite{scarpone2020multimethod,doblhammer2020social,miller2020mapping}. However, such maps suffer from the intrinsic limitation caused by the arbitrary boundaries imposed by the pre-defined administrative regions considered and exhibit undesirable discontinuities at these boundaries. Geostatistics provides the framework to model spatial variability and to account for data with varying spatial support delivering desegregated and continuous maps of the risk of disease, overcoming the limitations of choropleth maps. Additionally, geostatistical risk maps can integrate data uncertainty resulting from population size, which is very important in this case, given that the population density varies widely across the country. 

A number of methods that estimate the spatial risk of diseases from known datapoints have been proposed \cite{gatrell1996spatial}, including distance-based approaches and kernel densitiy estimation (KDE) methods. These and other methodologies may be used to convert point data to continuous risk maps. However, these methods have been designed to model static or slow-varying distributions, and cannot be trivially adapted to dynamic modeling of diseases.
Combined with time-series prediction, KDE has had some success in visualizing disease cluster evolution but depends critically on the accuracy of the time-series prediction method. 

Direct estimation of the geospatial risk of COVID-19 in Switzerland using high-resolution spatiotemporal data analysis
\cite{de2020geospatial} based on the modified space–time density-based spatial clustering of application
with noise (MST-DBSCAN) algorithm \cite{kuo2018characterizing} has been proposed. However, high spatiotemporal resolution health data is rarely available, due to limited information and other limiting factors, such as personal data protection policies.

Our modeling of the geospatial risk model, described in Section \ref{sec:dataset} is based on geostatistical block sequential simulation \cite{Oliveira2013stochastic}, assuming a Poisson model for rare diseases as proposed by Waller and Gotway \cite{waller2004applied} to estimate the local distributions functions \cite{Goovaerts2005}. Given the actual municipality-level data, with varying support, this method makes it possible to integrate this uncertainty in the spatial model to generate the incidence rate maps and to assess its spatial uncertainty at any region in the country. 

\subsection{Compartmental models}

Compartmental models have been extensively used in epidemiological studies. The first proposals to group populations in different compartments and to use differential equations to model the transitions between compartments, deriving the dynamics of the processes, date back more than a century \cite{ross1916application,kermack1927contribution}. Compartmental models use continuous differential equations to model the evolution of an epidemic in a population of individuals, where each individual is assigned a state (a compartment) that determines its stage in the epidemic process \cite{kermack1927contribution,bailey1975mathematical}. Since then, many articles have been published on the use of compartmental models to predict and simulate the dynamics of epidemics. In the last year alone, hundreds of articles have been published on the application of compartmental models to study the dynamics of the COVID-19 pandemic in many countries and locations, such as China, Italy, and India \cite{caccavo2020chinese,calafiore2020time,chatterjee2020studying}. A complete description of the many applications of the models, as well as the variants proposed, is outside the scope of this work.

Our own application of compartmental models to predict the dynamics of the pandemic in Portugal is based on the  Susceptible-Infected-Recovered-Dead (SIRD) model, which uses four different compartments. Despite the model simplicity, many assumptions and approximations are required to make it fit real-world data, including the consideration of time-varying parameters, adjustments for small-sample, high-variance, data, and estimation of unknown values in the observed data series. We used, with adaptations described in section \ref{sec:SIRD}, the methods proposed in recent work that used data from the COVID-19 pandemic in New York City, Madrid, and Stockholm, among other cities, together with various states, countries, and regions to estimate a SIRD model \cite{fernandez2020estimating}.

\subsection{Spatio-temporal modeling using neural networks}

Recurrent neural network (RNN) architectures, based on Long Short-Term Memory (LSTM) units \cite{schmidhuber1997long} or gated recurrent units (GRU) \cite{cho2014learning} have been extensively employed in forecasting tasks involving time-series data. However, these models consider the input data as independent sequences of vectors and do not explore the spatial context available in spatio-temporal data. 

To address this limitation, spatial convolution operations were combined with LSTMs, thereby exploiting the abilities of convolutional neural networks (CNNs) and RNNs to effectively model spatial and temporal information, respectively. In the  Convolutional LSTM (ConvLSTM) approach \cite{shi2015convolutional}, all input data structures are 3D tensors, with the first dimension corresponding to either the number of measurements or the number of feature maps, and the last two dimensions representing the spatial dimensions (i.e., width and height). By replacing the product operation in the original LSTM with the convolution operation, the future states of a certain cell depend on the inputs and past states of itself and its local neighbours. 

Many different architectures have been proposed using ConvLSTM building blocks, applied to different problems, such as weather prediction from satelite image sequences \cite{hong2017psique}, air quality forecasting \cite{alleon2020plumenet}, and video frame prediction \cite{wang2017predrnn}.
In the architecture that served as the basis for this work \cite{castro2021stconvs2s} the authors used a standard encoder-decoder structure comprised of multiple stacked ConvLSTM layers, with each decoder layer initializing its hidden states from the output of the corresponding encoder layer. The final predictions of the model are given by the concatenation of the
hidden states from the decoder network followed by a $1 \times 1$ convolution. This method was adapted to this problem by introducing a number of improvements, described in section \ref{sec:STS}.

\section{Computing incidence rate maps using block-DSS\label{sec:dataset}}

We use the incidence rate, the number of confirmed positive tests in a region, in a period of 14 consecutive days, divided by the population of the region, as a proxy for assessing the risk of infection. In particular, we consider the incidence rate per municipality \(z_\alpha(t)\), defined by
\begin{equation}
z_\alpha(t)=\frac{c_\alpha(t)}{n_\alpha}
\end{equation}
where \(c_\alpha(t)\) is the number of confirmed positive tests in the 14 days preceding a given day $t$, at each municipality, \(\alpha\),  and  \(n_\alpha\) is the respective population size. We will use $z_u$ to denote the incidence rate at a specific location $u$ (or cell) in the country. In the sequence, we may drop the explicit dependence on $t$ to simplify the notation, when not required. 

The daily numbers of new cases by municipality are reported regularly by the Portuguese Directorate-General for Health (DGS), and they have been converted to predicted incidence rate by dividing by the population count of each municipality, as given by official statistics, and scaled to 100,000 inhabitants (thus corresponding to the number of new cases in the most recent 14 days per 100,000 inhabitants). This will be the scale used in the maps and reports in this paper.

Since not all cases are detected, this incidence rate may underestimate the real incidence rate. However, it is reasonable to assume that the ratio between the actual number of newly infected persons and the number of detected cases is approximately equal in the different regions of the country, enabling us to use the incidence rate computed in this way as a proxy to the real incidence rate. In practice, we may be underestimating the incidence rate by a signficant factor, since it is well known that a relevant fraction of the COVID-19 cases are asymptomatic and remain undetected. This has an impact on the value of the parameters of the dynamic models, since important characteristics of the process such as the group immunity threshold, depend on the actual rates of infection and recovery. In practice, we believe this underestimation does not have a strong impact in the accuracy of the models (other than a systemic underestimation), since all model parameters are estimated from actual data and updated as time goes by in order to reflect the dynamics of the pandemic. Therefore, the only significant impact of the fact that not all cases are detected is the underestimation of the incidence rate, by a factor that can be considered approximately constant, if one assumes that there are no significant differences in testing strategy from region to region. This assumption can be justified by the relative homogeneity of the health system in a country like Portugal.

\begin{figure}[!htb]
\minipage{0.49\textwidth}
\includegraphics[width=\linewidth]{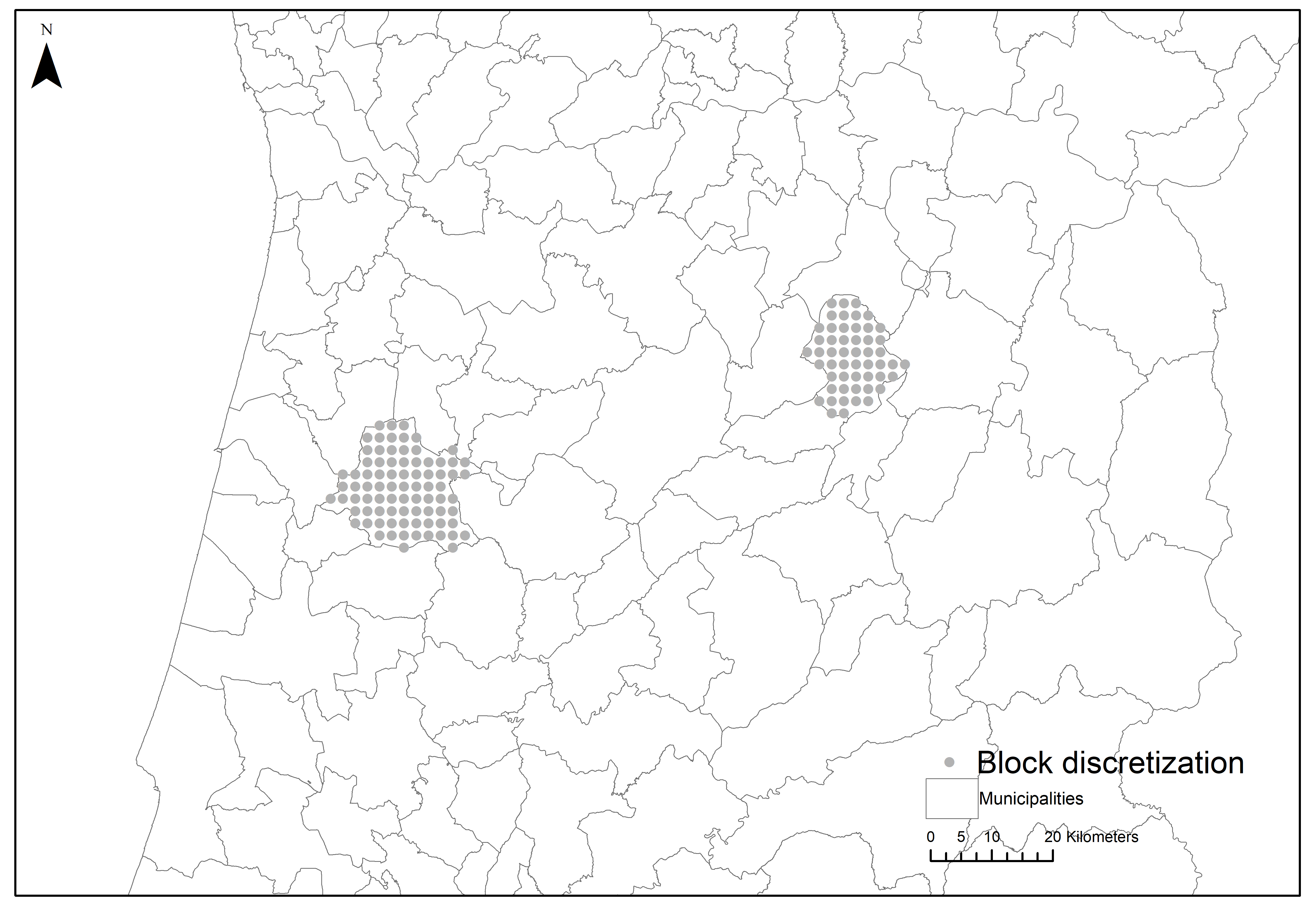}
\caption{\label{fig:cells} Example of discretization into cells, for two specific municipalities.}
\endminipage
\end{figure}

\begin{figure}[!htb]
\minipage{0.49\textwidth}
\begin{center}
\includegraphics[width=0.75\linewidth]{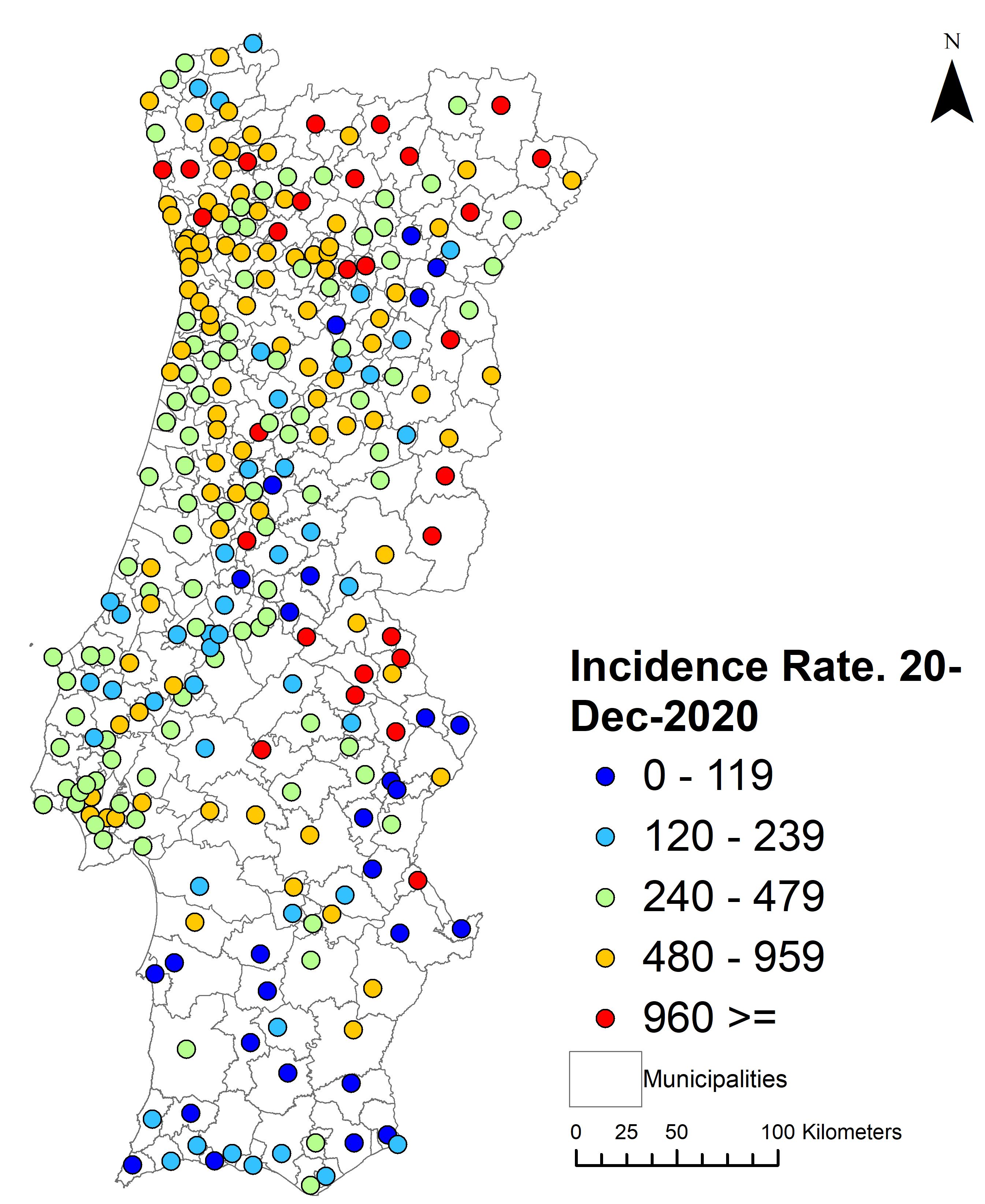}
\end{center}
\caption{\label{fig:municipalities} Municipalities in mainland Portugal, and corresponding centroids and incidence rates for a specific day in December 2020.}
\endminipage
\end{figure}

We consider the country divided in a regular grid of $N=40608$ square cells, which, in this work, have a dimension of 2 km by 2 km (see Figure \ref{fig:cells} for an example of the discretization used). 
\ref{fig:municipalities}.
The methodology can be trivially adapted to a different grid.  
We compute the incidence rate maps by estimating $z_u$ for each cell in the country, using geostatistical stochastic simulation, namely direct block sequential simulation (block-DSS) \cite{Liu09}, which is based on a stationary spatial covariance model (i.e., stationary variogram model). This modelling technique allows the daily update of COVID-19 incidence rate maps at high-resolution together with the associated spatial uncertainty. The geostatistical incidence rate maps are continuous (up to the scale of the discretization) and their spatial distribution follows a given spatial pattern as revealed by a variogram model inferred from the data. The incidence rate maps do not show any sharp discontinuities and are not influenced by administrative boundaries, such as municipality borders. Since incidence rates refer to different population sizes, these must be weighted when calculating the experimental variogram such that municipalities with large populations have greater weighting.

Below, we briefly describe this geostatistical simulation method. A full description of the method can be found in previous work by the authors \cite{Azevedo20}. 
We assume that the geometric centroid of each municipality \(\alpha\) has coordinates \((x_\alpha,y_\alpha)\). Data is available for the 278 municipalities in mainland Portugal, as illustrated in Figure \ref{fig:municipalities}.
For clarity, in this section we will drop the explicit dependency on $t$. The daily update of incidence rates \(z_\alpha\) known for each municipality are assigned to each centroid (weighted by population density) and are used as experimental data in the geostatistical simulation. This is a reasonable approximation given the relatively small area of each municipality when compared with the area of the country.

Within a geostatistical framework \(z_\alpha\) is interpreted as a realization of a random variable \(Z_\alpha\) corresponding to the true, and unknown, incidence rate in municipality $\alpha$. The expected value of \(E[Z_\alpha]\) is an approximation of the incidence rate. \(Z\) depends on the population size of each municipality. For example, a given municipality with small population size $n_\alpha$ (i.e., a small denominator) will have high variance and consequently higher uncertainty in the incidence rate estimate. We use the Poisson kriging model \cite{Goovaerts2005} to define the risk variance
\begin{equation}
Var[Z_\alpha]=\sigma_R^2+\frac{E[Z_\alpha]}{n_\alpha}.
\end{equation}
Block-DSS \cite{Liu09} is a stochastic sequential simulation method based on the Poisson kriging model and an adequate modelling technique to deal with data with varying spatial support (i.e., municipalities with varying size and shape) and to make spatial predictions with change of support, e.g., from area (block) to point support. In this case the scale related to the map cells can be referred as point support, because it denotes a small area when compared with the municipality areas. The following sequence of steps summarizes the block-DSS algorithm \cite{Liu09}:
\begin{enumerate}
    \item Generate a random path that visits each cell, \({u}\), of the simulation grid;
	\item At each location along the random path, \({u}\), search the conditioning data within a given neighborhood dependent on the variogram model. The conditioning data comprises the closest experimental data, previously simulated values and block data;
	\item Calculate the local covariance values considering spatial covariance matrices between: block-to-block, block-to-point, point-to-block and point-to-point. The point data represents the incidence rates assigned at the centroid of each municipality and the block data are defined as the spatial linear average of point values. These matrices are built to solve the block kriging system and obtain the local mean and variance kriging estimate at location \({u}\) \cite{Liu09};
	\item Draw a value, $z_u$, from the global probability distribution function centered at the local mean and bounded by the local variance obtained in (3);
	\item Add the simulated value to the data set and repeat steps (3) to (6) until all grid cells are simulated for one realization;
\end{enumerate}

\begin{figure}[!htb]
\minipage{0.49\textwidth}
\begin{center}
\includegraphics[width=0.75\linewidth]{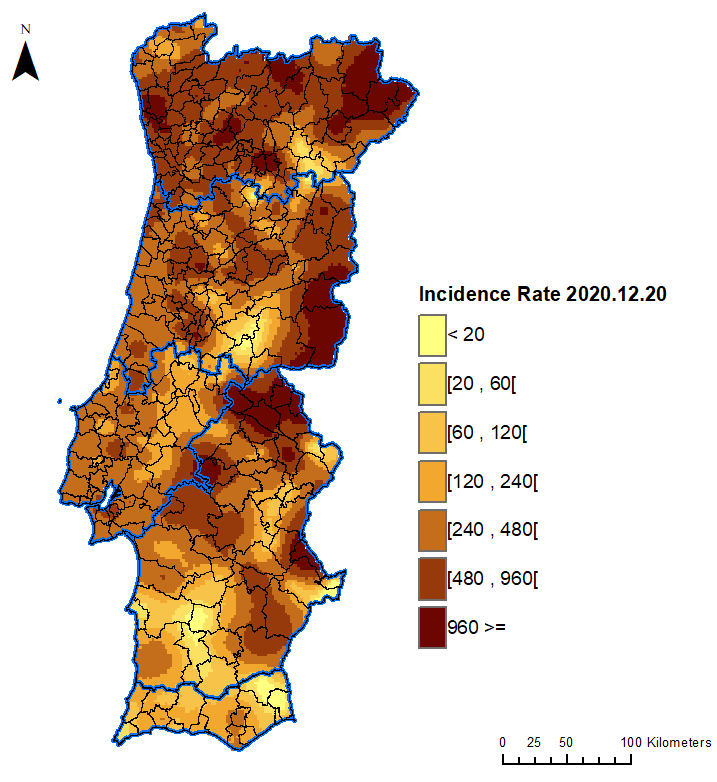}
\end{center}
\caption{\label{fig:risk}Computed incidence risk for December 20th, 2020.}
\endminipage
\end{figure}

\begin{figure}[!htb]
\minipage{0.49\textwidth}
\begin{center}
\includegraphics[width=0.75\linewidth]{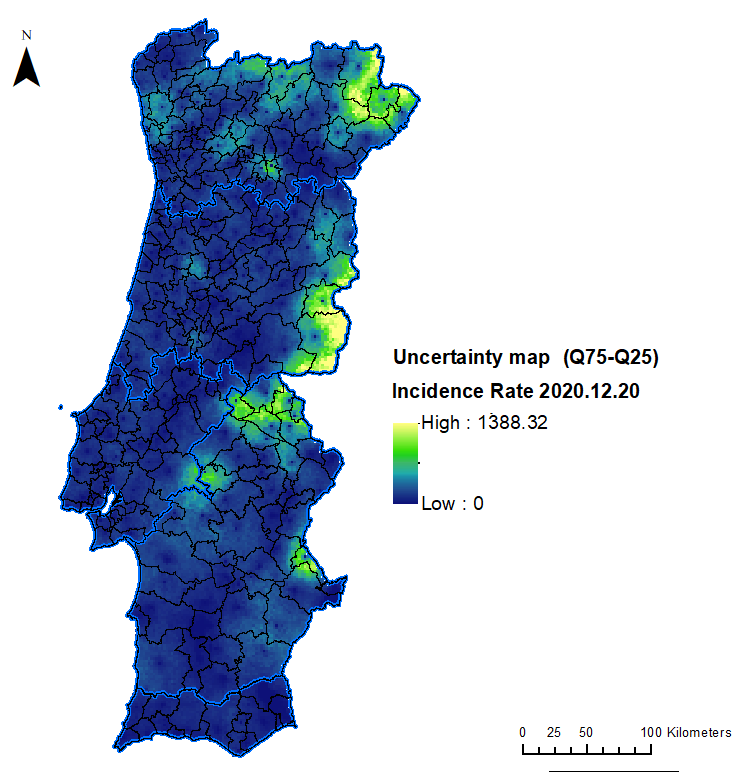}
\end{center}
\caption{\label{fig:risk_ci}Computed uncertainty maps for December 20th, 2020.}
\endminipage
\end{figure}

We applied the algorithm to compute the incidence values for each cell in the territory, for each day in the period under consideration. Figure \ref{fig:risk} depicts an example of this computation, for a specific day in the period.
Block-DSS generates alternative incidence rate maps, designated realizations, at each run, as the random path changes and consequently the conditioning data when simulating a location \({u}\). A given set of realizations provides the value of the uncertainty of incidence rate distribution in a given cell. In this work we simulate daily sets of one hundred realizations of incidence rates for mainland Portugal. Then, we computed the median incidence rate model as it represents the most likely incidence rate at a given period of time and location. The method also derives confidence intervals for the incidence rate in each cell (see example in Figure \ref{fig:risk_ci}). 
The value of the incidence rate at each cell represents a good approximation, given existing data limitations, to the real incidence rate at that location. Both the incidence risk maps and the confidence intervals are made available as supplemental data.

\section{Modeling methods\label{sec:methods}}
To model the spatio-temporal spread of the COVID-19 epidemic in Portugal, we implemented four models: 
\begin{itemize}
    \item an ARMA autoregressive-moving-average at the cell level;
    \item a vector-autoregressive model at the cell level that takes into account the previous values of all other cells;
    \item a municipality level compartmental SIRD model coupled with block-DSS kriging;
    \item a spatio-temporal convolutional sequence-to-sequence neural network.
\end{itemize}
All models produce predictions at the cell level. These predictions are then compared with the gold standard obtained using the method described in section \ref{sec:dataset}.

\subsection{Autoregressive moving average model \label{sec:ARMA}}
The baseline ARMA(p,q) model is a simple, cell-level, autoregressive-moving-average model \cite{whittle1963prediction}. In this model, the incidence rate for each cell $u$ is computed using 
\begin{equation}
z_u(t) = k_u + \epsilon_u(t)+\sum_{i=1}^p \varphi_u(i) z_u(t-i)+\sum_{i=1}^q \theta_u(i) \epsilon_u(t-i)
\end{equation}
where $k_u$, \(\varphi_u(i)\) and \(\theta_u(i)\) are calculated, for each cell $u$, using linear regression, in order to minimize the observed errors, \(\epsilon_u(t)\).

We used the ARIMA package in Python to compute an independent ARMA model for each cell $u$. The resulting model was used to predict the temporal evolution of the incidence rate $z_u(t)$ for each cell $u$ at time $t$.

\subsection{Vector autoregressive model \label{sec:VAR}}
The ARMA model suffers from a strong limitation, since it computes the evolution of the incidence rate at each cell not taking into account the values of any other cells. To circumvent this limitation we also tested a  vector autoregressive (VAR) model \cite{sims1980macroeconomics}, which computes the incidence rate at each cell using all the values from the other cells, in accordance with 
\begin{equation}
% Z(t) = K + \epsilon(t)+\sum_{i=1}^p A(i) Z(t-i)
z_u(t) = k_u + \epsilon_u(t) + \sum_{i=1}^p \sum_{v=1}^N A_{u,v} (i) z_v(t-i)
\end{equation}
where $z_u$ denotes the incidence rate and $A$ is an $N \times N$ time-invariant matrix, calculated, together with $k_u$,  using linear regression, in order to minimize the observed errors, \(\epsilon_u(t)\). 
We used the VAR package in Python to compute a VAR model for the data and a prediction of $z_u(t)$, for each cell $u$ at time $t$.

\subsection{Compartmental models coupled with block-DDS\label{sec:SIRD}}

\subsubsection{SIRD}

In the  SIRD compartmental model used at the municipality level, a given person can be in four different states, in what regards his or her relation with the disease: susceptible (S), infected (I), recovered (R), and dead (D).
Under this model, the dynamics of the epidemic are then modeled using the following system of differential equations
\begin{equation}
\frac{d S}{d t} = - \frac{\beta I S}{N}, 
\label{eq:suc}
\end{equation}
\begin{equation}
\frac{d I}{d t} = \frac{\beta I S}{N}-\gamma I-\delta I,
\label{eq:infec}
\end{equation}
\begin{equation}
\frac{d R}{d t} = \gamma I,
\label{eq:recov}
\end{equation}
\begin{equation}
\frac{d D}{d t} = \delta I, 
\label{eq:dead}
\end{equation}
where \(\beta\), \(\gamma\), and \(\delta\) are the rates of infection, recovery and mortality, respectively.

We computed a separate SIRD model for each municipality, $\alpha$, with independent parameters \(\beta_\alpha\), \(\gamma_\alpha\), and \(\delta_\alpha\), for a total of 278 municipalities. Furthermore, as other authors have done \cite{fernandez2020estimating}, we removed the assumption of the basic SIRD model that the rates \(\beta\), \(\gamma\), and \(\delta\) are constant and depend only on the nature of the epidemic and the fixed spreading dynamics. Since, in practice, the behavior of an epidemic is sensitive to changes in human behavior and to other environmental variables, we used a generalized version of the SIRD model that uses time-varying rates, which are calculated from known pandemic data, at each municipality \(\alpha\), resulting in a series of values for each parameter:

\begin{equation}
    \beta_\alpha(t)=\frac{i_\alpha(t) }{I_\alpha(t)}\\
    \label{eq:params_calc1}
\end{equation}
\begin{equation}
    \gamma_\alpha(t)=\frac{r_\alpha(t)}{I_\alpha(t)}\\
    \label{eq:params_calc2}
\end{equation}
\begin{equation}
    \delta_\alpha(t)=\frac{d_\alpha(t)}{I_\alpha(t)}
    \label{eq:params_calc3}
\end{equation}
where \(i_\alpha(t)\) represents the new confirmed cases at time \(t\), \(r_\alpha(t) = R_\alpha(t)-R_\alpha(t-1)\) the daily recovered,  \(d_\alpha(t) = D_\alpha(t)-D_\alpha(t-1)\) the daily deaths, and \(I_\alpha(t)\) the total active cases at time \(t\), given by
\begin{equation}
I_\alpha(t_f)=\sum_{t=t_0}^{t_f} i_\alpha(t)-[r_\alpha(t)+d_\alpha(t)].
\label{eq:activeCases}
\end{equation}

To avoid instability and to reflect the fact that the \(\beta\), \(\gamma\), and \(\delta\) rates vary slowly, once the parameters are calculated for the entire historical data, kernel smoothing is applied to reduce noise. We use Nadaraya-Watson as the kernel with a bandwidth of 10.

\subsubsection{Forecasting}
Each Portuguese municipality \(\alpha\) is treated as an independent region, modeled by its own set of SIRD equations, which can be used to forecast the evolution of the cases. In order to predict the future number of infections for each municipality, we solve the SIRD model forward in time using the last day of historical data as the initial conditions for $S$, $I$, $R$ and $D$. We obtain the values for \(\beta_\alpha(t)\), \(\gamma_\alpha(t)\) and \(\delta_\alpha(t)\) to use during the forecast period based on the historical smoothed parameter series. This can be done in three ways:  
\begin{enumerate}
  \item Using their last known value;
  \item Using an average of the last $n$ days;
  \item Estimating via extrapolation.
\end{enumerate}
\subsubsection{Pseudo-count Component}
Equations \ref{eq:params_calc1}, \ref{eq:params_calc2}, and \ref{eq:params_calc3} compute all parameters relative to the number of active cases. This can lead to severe fluctuations of parameter values when the population of the municipality is small, since the variance of the parameters will be high.
We mitigate this by introducing a pseudo-count component that forces parameter values to consider not only regional but national data as well - therefore avoiding unfounded radical shifts.

Let  $\beta_\alpha(t)$,$ \gamma_\alpha(t)$ and $\delta_\alpha(t)$ be the parameters for municipality $\alpha$ at time \(t\), and $\beta_P(t)$, $\gamma_P(t)$, $\delta_P(t)$ the values for the rates calculated on the national level. We can now calculate $\beta^{*}_\alpha(t)$ as
\begin{equation}
 \beta^{*}_\alpha(t)  = \frac{K \beta_P(t) + n_\alpha \beta_\alpha(t)}{K+n_\alpha},  
\end{equation}
where $K$ is the pseudo-count. If $K=0$, then the national parameters are ignored. As $K$ increases so does their influence but remains inversely proportional to the region's population size. The values for $\gamma_\alpha^{*}(t)$, $\delta_\alpha^{*}(t)$ are computed in the same way. \(\beta_\alpha^{*}(t)\), $\gamma_\alpha^{*}(t)$, and $\delta_\alpha^{*}(t)$ are then used in equations \ref{eq:suc} to \ref{eq:dead} to model the evolution of the relevant variables, instead of using the rates computed directly from the historical time series.

\subsubsection{Approximating the number of recoveries and deaths.}
Privacy laws create a challenge in obtaining the number of deaths for each region daily. Furthermore, it is difficult to determine the exact recovery date for every single patient accurately. As a way of overcoming these limitations, we approximate the values for each region's number of recoveries and deaths based on its population size, the mean recovery period, and the national mortality ratio. We used a mean recovery period of 14 days, which led to the best fit of the models to the real data. 

First, we calculate the national mortality rate for each day \(t\):
\begin{equation}
    \mu(t) = \frac{d_{P}(t)}{I_{P}(t-w)},
\end{equation}
where \(t\) is the current day, \(w\) is the expected recovery time, \(d_{P}(t)\) is the number of deaths in the country on day \(t\), and  \(I_{P}(t - w)\) is the total number of infected people in the country at time $t-w$.

Then, for each municipality \(\alpha\) and for each day \(t\), we estimate the number of recoveries and deaths, as follows:
\begin{equation}
    d_\alpha(t) = \mu(t) I_\alpha(t-w),
\end{equation}
\begin{equation}
    r_\alpha(t) = I_\alpha(t) - d_\alpha(t),
\end{equation}
where \(r_\alpha(t)\), \(I_\alpha(t)\), \(d_\alpha(t)\) are the number of recoveries, total active cases and deaths at time $t$, respectively.

\subsubsection{Computing the geographical incidence rate using the SIRD model\label{sec:predictSIRD}}

By solving the SIRD equations forward in time, with the estimated parameters, smoothed using pseudo-counts, we obtain the values of \(I_\alpha(t)\) for each municipality \(\alpha\) and each time $t$. From \(I_\alpha(t)\) we can trivially compute the number of new cases, for each day, which makes it possible to compute the incidence rate \(c_\alpha(t)\) for each time $t$. This data, generated by the SIRD model, is then used instead of the real data, as input to block-DSS algorithm, using the procedure described in section \ref{sec:dataset} to compute the incidence rate in each cell $u$ of the territory at time $t$, $z_u(t)$.

\subsection{Spatio-temporal convolutional sequence-to-sequence neural networks\label{sec:STS}}

\begin{figure*}[ht]
\centering
\includegraphics[width=\linewidth]{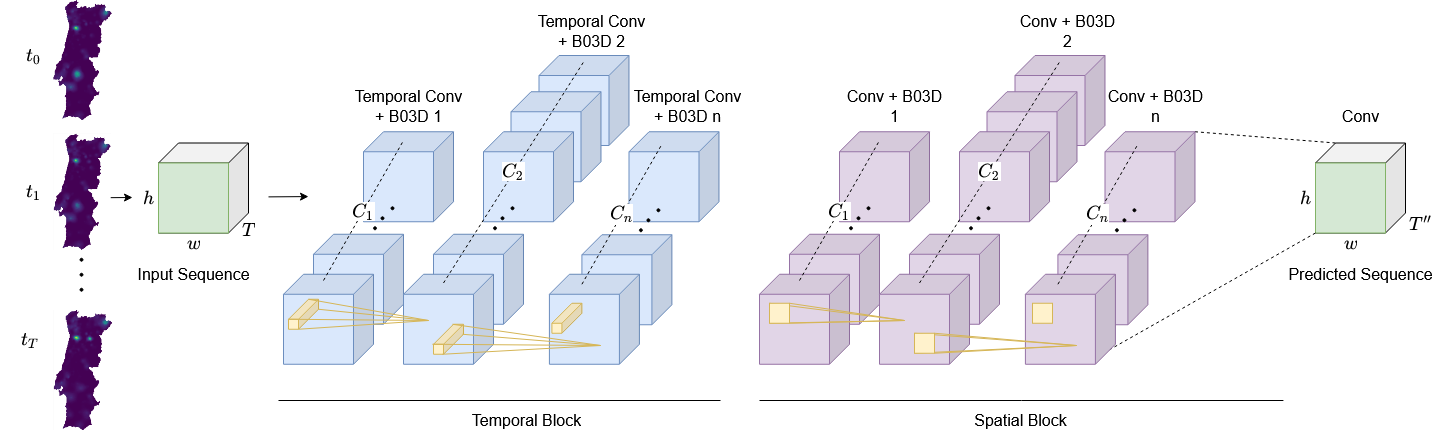}
\caption{The STConvS2S architecture \label{fig:STS}}
\end{figure*}

This approach uses recently proposed techniques for spatio-temporal modeling using neural networks, resulting in a model derived from the STConvS2S architecture \cite{castro2021stconvs2s}, coupled with new normalization-activation layers and learnable parameters intrinsic to each location. 

The STConvS2S architecture is a sequence-to-sequence model comprised exclusively of convolutional layers, which are suited to capture spatial features by design (see Figure \ref{fig:STS}). Convolutions are performed with factorized 3D kernels $K = t\times d\times d$, where $t$ is the size of the temporal kernel and $d$ is the size of the spatial kernel. 

The first component is a temporal block that extracts temporal features through the temporal kernels (i.e., $t\times 1\times 1$), while the second component is a spatial block that receives the output of the previous block in order to extract spatial features through the spatial kernels (i.e., $1\times d\times d$). Both the temporal and spatial blocks are comprised of successive convolutional blocks with normalization-activation operations, with the temporal block using causal convolutions to maintain temporal coherency during prediction (i.e., predictions for a time-step $t$ make no use of future information from time-steps $t + 1$ onward). 

The feature maps generated throughout the architecture are adequately padded in order to maintain the initial dimensions, and the intermediate feature maps are increased by a factor of 2 in each layer, with the last convolutional layer in both the spatial and temporal block reducing the number back to the initial amount.

Instead of the traditional batch normalization and activation function operations following each convolution, we apply an extension of the normalization-activation layers recently proposed \cite{liu2020evolving}. The chosen operation is an adapted version of the EvoNormB0 layer, which was the best performing batch-based version reported, and explicitly models spatio-temporal scenarios, considering sequences of two-dimensional inputs when calculating both the batch and instance variance (i.e., the values associated to each time-step in the input sequence are considered separately when computing the variance). Let $v_1$, $\theta$ and $\psi$ denote learnable parameter vectors, and let $s_{b,d,w,h}$ and $s_{d,w,h}$ represent the variance of a mini-batch and the variance of a single instance, respectively. This extension, denoted here as B03D, is defined as follows:

\begin{equation}
    \mathrm{B03D} = \dfrac{x}{\mathrm{max}\left(\sqrt{s_{b,d,w,h}^2\left(x\right) + \epsilon}, v_1 \cdot x + \sqrt{s_{d,w,h}^2\left(x\right) + \epsilon}\right)}\cdot \theta + \psi
\end{equation}

Another technique used in our model was the extension of the input data with learnable local features and weights, intrinsic to each spatial location, as originally proposed for the area of wind forecasting, where changes in location imply changes in the model parameters \cite{uselis2020localized}. This technique seeks to improve spatio-temporal forecasting scenarios by combining both (i) global location-invariant features, and (ii) location-specific features. 

Typical CNNs are mostly translation invariant. Translating an input $x$ and convolving with a filter $k$ will yield the same result as translating the feature map resulting from a convolution between $x$ and $k$. Therefore, standard CNNs treat each spatial location equally and thus learn global patterns. This is often insufficient in many spatio-temporal forecasting scenarios, where the behaviour in specific locations should be guided by their own intrinsic local features. 

To allow the learning of such features, we used two complementary techniques, Learnable Inputs (LI) and Local Weights (LW). The LIs correspond to a set of $n$ trainable parameters with the same spatial dimensions as the original input, concatenated with the input before processing with a convolutional layer. These parameters allow the network to learn local features regarding every spatial location, which can complement the learned global patterns or be ignored by the convolutional kernels in situations where they are not beneficial. The LWs further reinforce the individualized learning of different spatial locations, through a locally-connected layer (i.e., a convolutional layer with a different filter at each input region, allowing different spatial locations to be weighed according to their relevance) of weights over the input, resulting in $m$ trainable weights with the same spatial dimensions as the input. Similar to LIs, these weights are afterwards concatenated with the original input. In our experiments, both the LIs and LWs are concatenated with the inputs on the channel dimension (sharing the LIs across the different input time-steps and using different LWs in each time-step), prior to every convolutional layer. LIs are implemented with a locally connected layer that receives as input a constant unitary tensor with the same spatial dimensions as the original input, and the LWs are implemented with a separate locally connected layer that receives the original tensor as input.

\section{Experimental comparison of modeling accuracy\label{sec:results}}

\begin{figure}[!htb]
\minipage{0.49\textwidth}
\includegraphics[width=\linewidth]{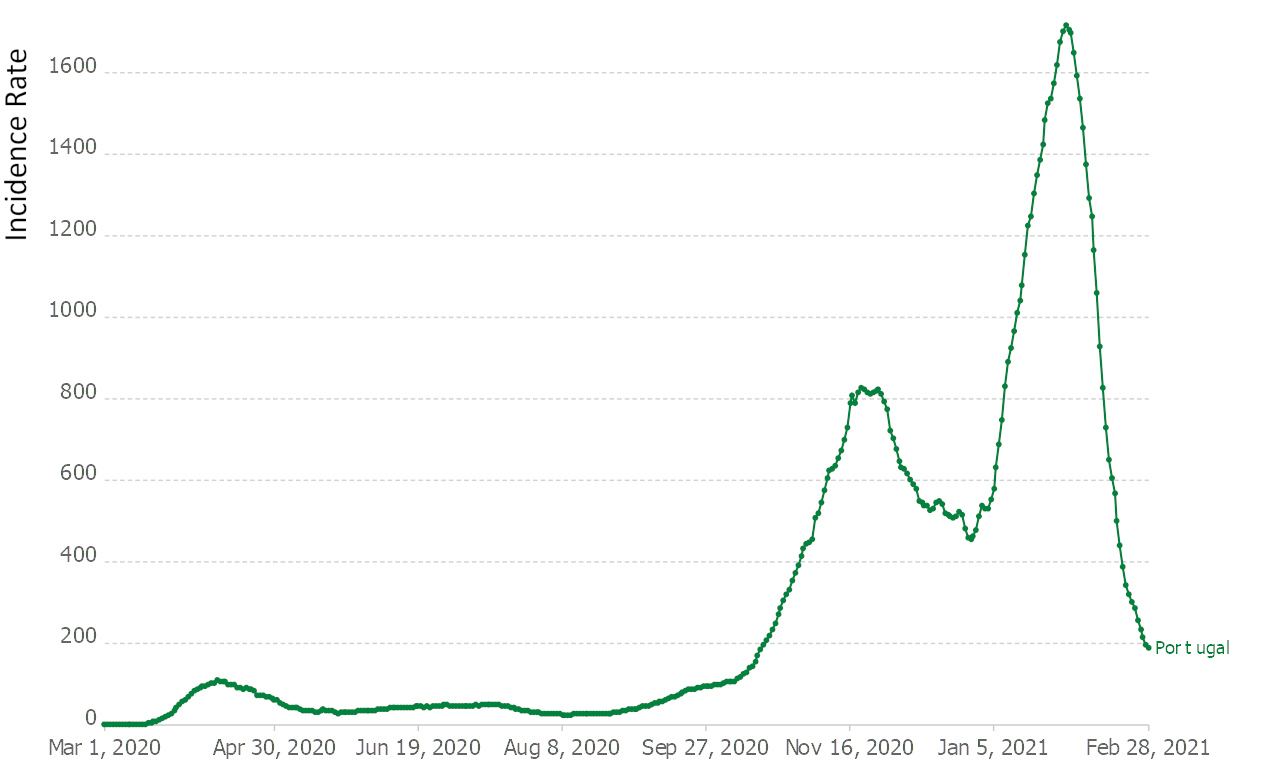}
\caption{\label{fig:incidence_rate_portugal} Evolution of the incidence rate in Portugal, for the first year of the COVID-19 pandemic.}
\endminipage
\end{figure}

We performed an empirical comparison of the four methods described in section \ref{sec:methods} by using them to predict the incidence rate $7$ and $10$ days ahead (corresponding to time-step $t+T$, with $T=7$ and $T=10$), using only past data up to time-step $t$. The methods were used to predict this variable from September 7th, 2020, until February 28th, 2021. Each model was initially trained with data from the first six months of the pandemic (March 1st 2020 - August 31st 2020) and, as the prediction moved forward, was allowed to use existing data up to $T$ days before the day being predicted. During the periods used for training and testing (twelve months) Portugal faced three major waves of the infection. The prediction period included the two largest waves, which reached incidence rates never seen during the training period. Figure \ref{fig:incidence_rate_portugal} shows the 14-day incidence rates, per 100,000 inhabitants, for the whole country, during the first year of the pandemic.

\subsection{Model training and parameterization}

The ARMA and VAR baseline models used as input data the entire past sequence for each cell up to time $t$. They were then used to predict the $T$ next days sequentially. We considered only the last value in the output sequence, corresponding to the time-step $t+T$, and the process was repeated by sliding the window one day into the future and concatenating the ground-truth value corresponding to the time-step $t+1$ with the existing data.

For the SIRD compartmental models no training is necessary. As explained in section \ref{sec:predictSIRD}, the time-varying parameters for each municipality were determined and used to solve the equations forward in time. For each parameter we calculate each day in the historical dataset and then use extrapolation to project its values forward $T$ days. By solving the equations for each municipality $\alpha$ we get a prediction of the future number of new cases for that municipality, $z_{\alpha}$. We then applied the block-DSS algorithm to determine the incidence rate for each cell in the territory, as described in section \ref{sec:predictSIRD}. The application of the block-DSS algorithm used the same spatial covariance matrix as the one used to build the gold standard data set.   

For the STConvS2S architecture, the model received as input a sequence of $T$ contiguous days, and generated a sequence corresponding to the next $T$ days. The final prediction is thus given by considering only the last time-step in the output sequence. This process was repeated by sliding the input window one day into the future, as was done for the ARMA and VAR models. Furthermore, in order to incrementally update the parameters as new data becomes available, we apply a simple online learning procedure, fine-tuning the model with each new input sequence for $5$ epochs, after the prediction was made. This enables the model to quickly adapt to sudden changes.

Regarding hyperparameter choices, for the ARMA model we consider $p = 7$ and $q = 1$. For the VAR model, we used $p=4$. For the SIRD model, we used the pseudo-count parameters $K=10,000$ in the prediction 7 days ahead and $K=100,000$ in the prediction 10 days ahead. This values were selected as the one that obtained better results. As in the original proposal, the STConvS2S architecture used $3$ convolutional layers for both the temporal and spatial block, plus the final convolution to reduce the channel dimensionality back to one. The initial number of convolutional filters was set to $32$, and each filter was of dimensionality $5\times 5$. In order to select the most optimal Local Inputs and Learnable Weights configuration, we varied the filter size $k$ of the LWs between $1\times 1\times T$, $2\times 2\times T$ and $1\times 1\times 1$ (i..e, in this last case, considering a direct element-wise weighing unique to each spatial location and time-step in the input sequence), where $T$ corresponds to the number of time-steps in the input sequence. Furthermore, we varied the number of LIs and LWs $n \in \{1, 2, 3\}$. The parameters were finally set as $k = 1\times 1\times 1$ and $n = 2$. Training relied on the AdaMod optimizer using a learning rate of $10^{-3}$ and batch size of $5$. Furthermore, for the STConvS2S model, in order to fit in the GPU memory and to avoid losing information over individual regions, the same architecture was trained and tested on three different subsets of the input region, corresponding to the upper third, middle third, and lower third of Portugal. The final prediction results were then  concatenated, resulting in a prediction encompassing all of mainland Portugal.

\subsection{Experimental results}
The predictions made by each method were compared with the reference incidence rate values, obtained as described in section \ref{sec:dataset}. We computed two commonly used figures of merit in order to compare the predictions of the models with the reference data, the RMSE and the sMAPE.

For a given day $t$, the root mean square error (RMSE) between the  predicted value and the reference value, averaged over all cells, is given by:
\begin{equation}
    RMSE = \sqrt{\frac{\sum_{u} (z_u(t) - \widehat{z_u}(t))^2}{N}},
\end{equation}
where \(z_u(t)\) is, as before, the incidence rate in cell $u$ at time $t$ and \(\widehat{z_u(t)}\) is the value predicted by the model.

Since absolute values of the incidence rate vary greatly by location, it is also relevant to compute a second figure of merit, the symmetric mean average error (sMAPE):
\begin{equation}
    sMAPE = \frac{1}{N} \sum_u \frac{2 |z_u(t) - \widehat{z_u}(t)|}{z_u(t)+\widehat{z_u}(t)}.
\end{equation}

\begin{figure}[!htb]
\minipage{0.49\textwidth}
\includegraphics[width=\linewidth]{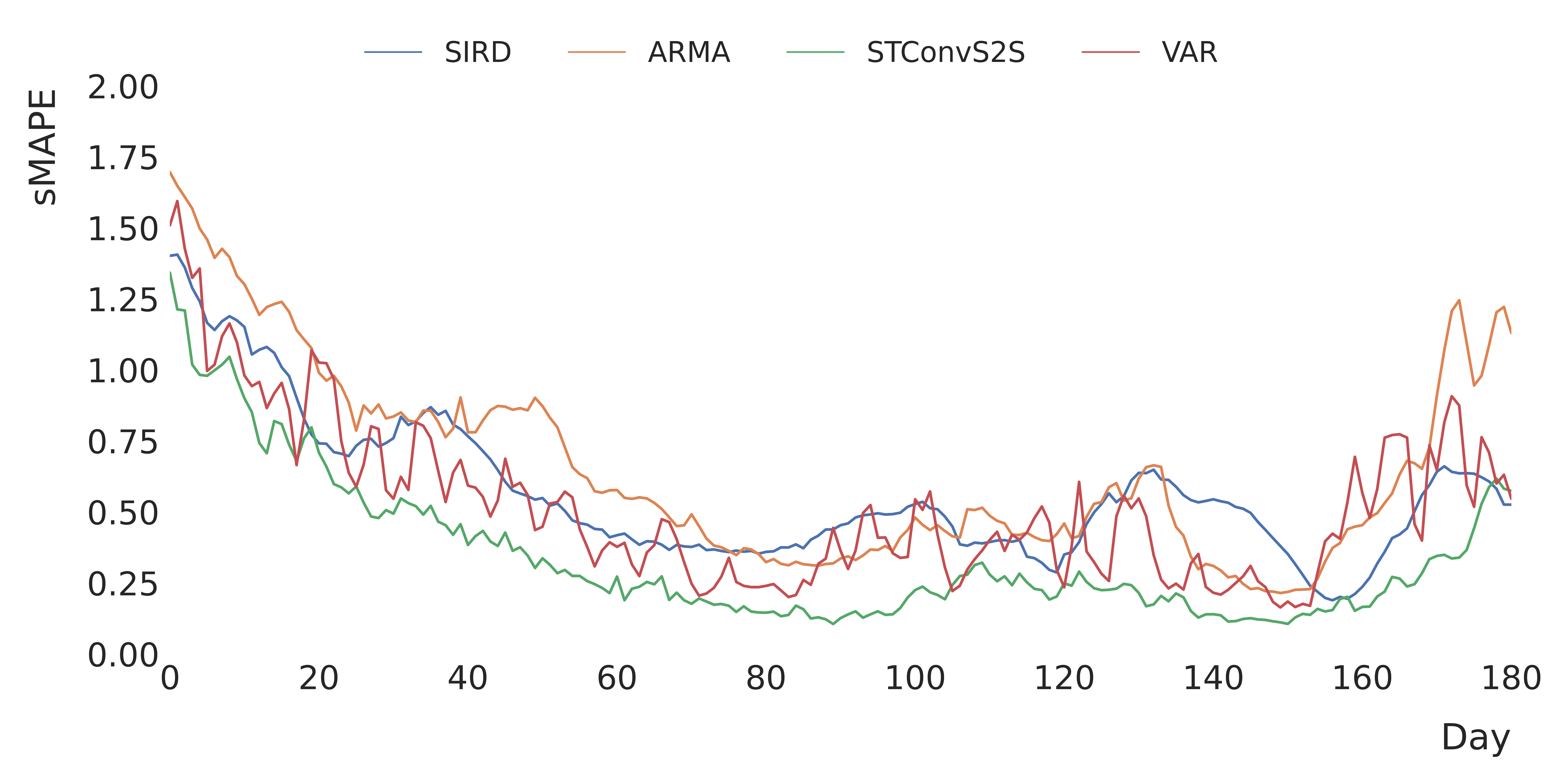}
\caption{\label{fig:SMAPE} Daily sMAPE for a prediction one week ahead.}

\endminipage\hfill
\minipage{0.49\textwidth}
\includegraphics[width=\linewidth]{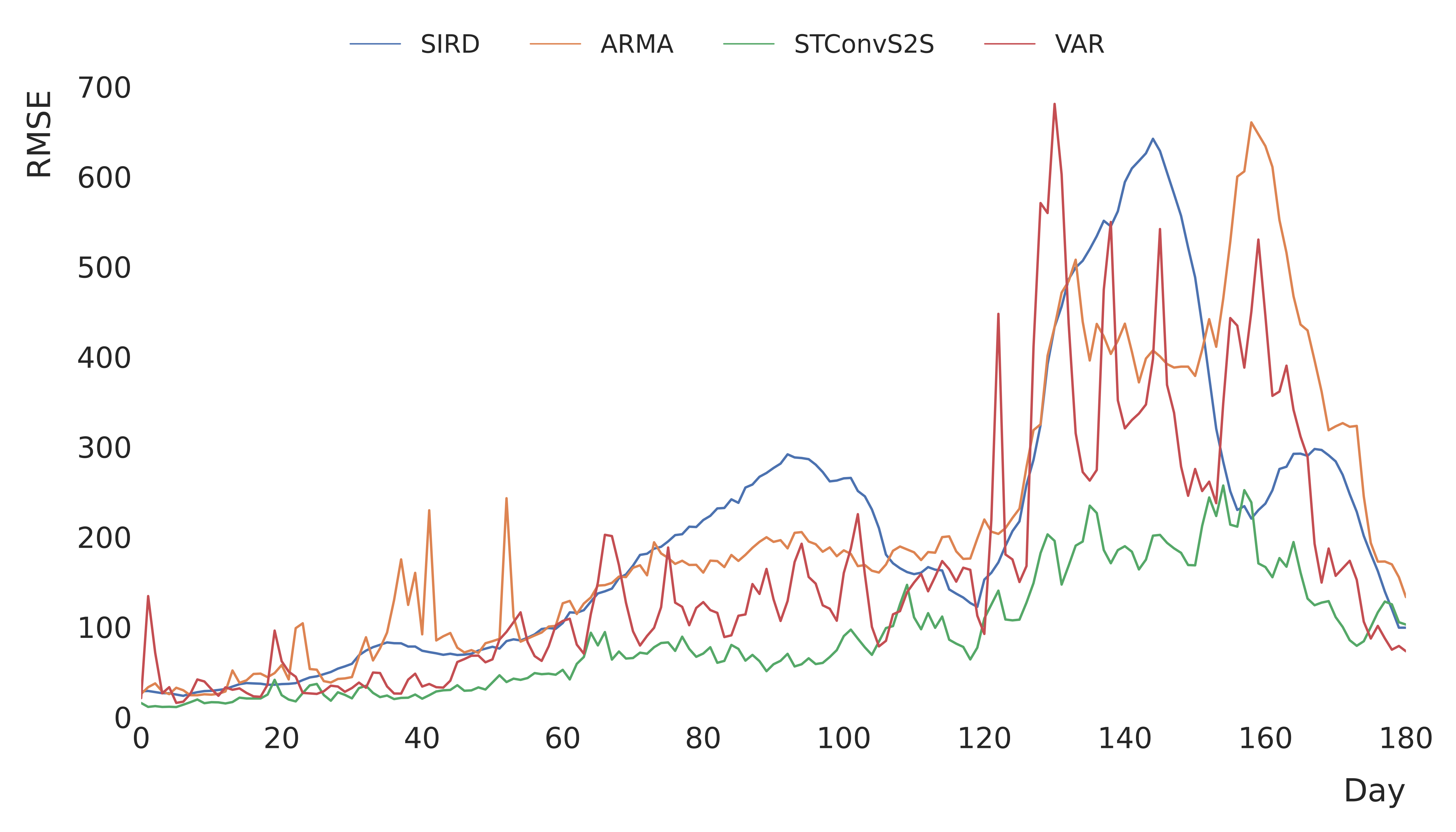}
\caption{\label{fig:RMSE} Daily RMSE for a prediction one week ahead.}
\endminipage
\end{figure}

Figure \ref{fig:SMAPE} and \ref{fig:RMSE} show the evolution of these two figures of merit for the four methods used (day 0 corresponds to September 7th, 202o). The superior performance of the STConvS2S method is clear, when compared with both the baseline ARMA and VAR models, and the SIRD model. As always, the incidence rate is reported as the number of new cases in the last 14 days per 100,000 inhabitants, and the value of the RMSE uses the same scale.

\begin{table}[htbp]
\begin{small}
\begin{tabular}{|l|r|r|r|r|}
\hline
       & \multicolumn{2}{c|}{7 days ahead ($\mu \pm \sigma$)} & \multicolumn{2}{c|}{10 days ahead ($\mu \pm \sigma$)}\\
\hline
Method & RMSE & sMAPE & RMSE & sMAPE\\
\hline
ARMA & $210.7 \pm 152.6$ & $0.653 \pm 0.345$ & $419.4 \pm 668.2$ & $0.817 \pm 0.365$\\
VAR & $162.6 \pm 139.7$ & $0.519 \pm 0.281$ & $230.9  \pm 210.3$ & $0.652 \pm 0.305$\\
SIRD & $204.0 \pm 156.0$ & $0.566 \pm 0.251$ & $195.0 \pm 152.6$ & $0.575 \pm 0.297$\\
STConv & $89.4 \pm 67.1$ & $0.342 \pm 0.252$ & $87.5 \pm 69.9$ & $0.331 \pm 0.250$\\
\hline
\end{tabular}
\end{small}
\caption{Average and standard deviation of RMSE and sMAPE, computed for all cells and all days in the period Sep. 8th 2020 - Feb 28th. 2021 \label{tab:results}, for prediction 7 and 10 days ahead.}
\end{table}
Table \ref{tab:results} shows the values for the RMSE and the sMAPE of the predictions, averaged over all days in the period September 7th, 2020, to February 28th, 2021. This table also shows the clear superiority of the STConvS2S algorithm. The SIRD model seems to be more stable as the horizon of the prediction is increased, although we needed to adjust the value of the hyperparameter $k$, so that the longer-term prediction uses more heavily the national level data in the computation of the parameters of the dynamics.

\begin{figure}[ht]
\centering
\includegraphics[width=\linewidth]{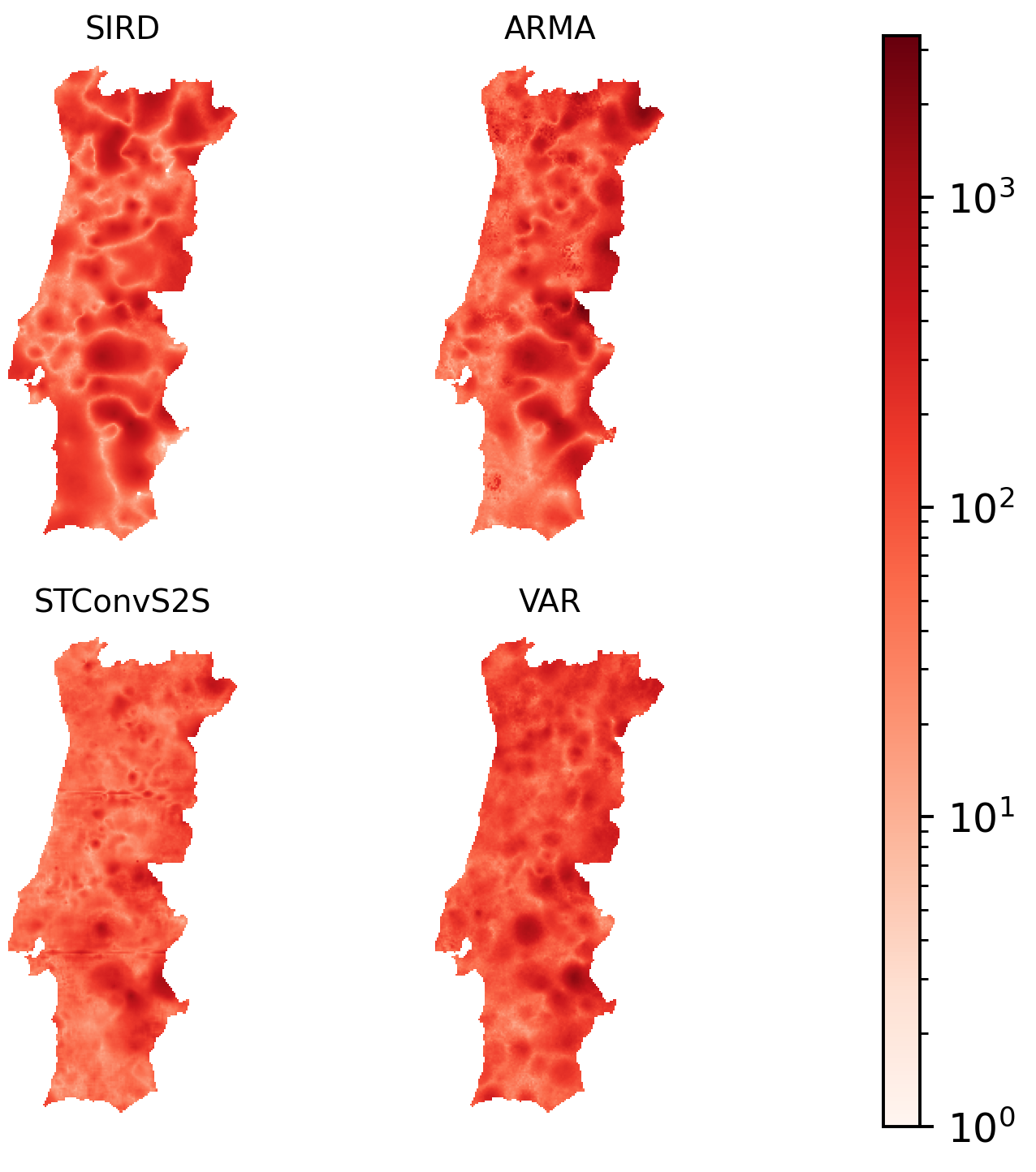}
\caption{Prediction error (RMSE) for the VAR, ARMA, SIRD and STConvS2S models. \label{fig:heatmap1}}
\end{figure}

\begin{figure}[ht]
\centering
\includegraphics[width=\linewidth]{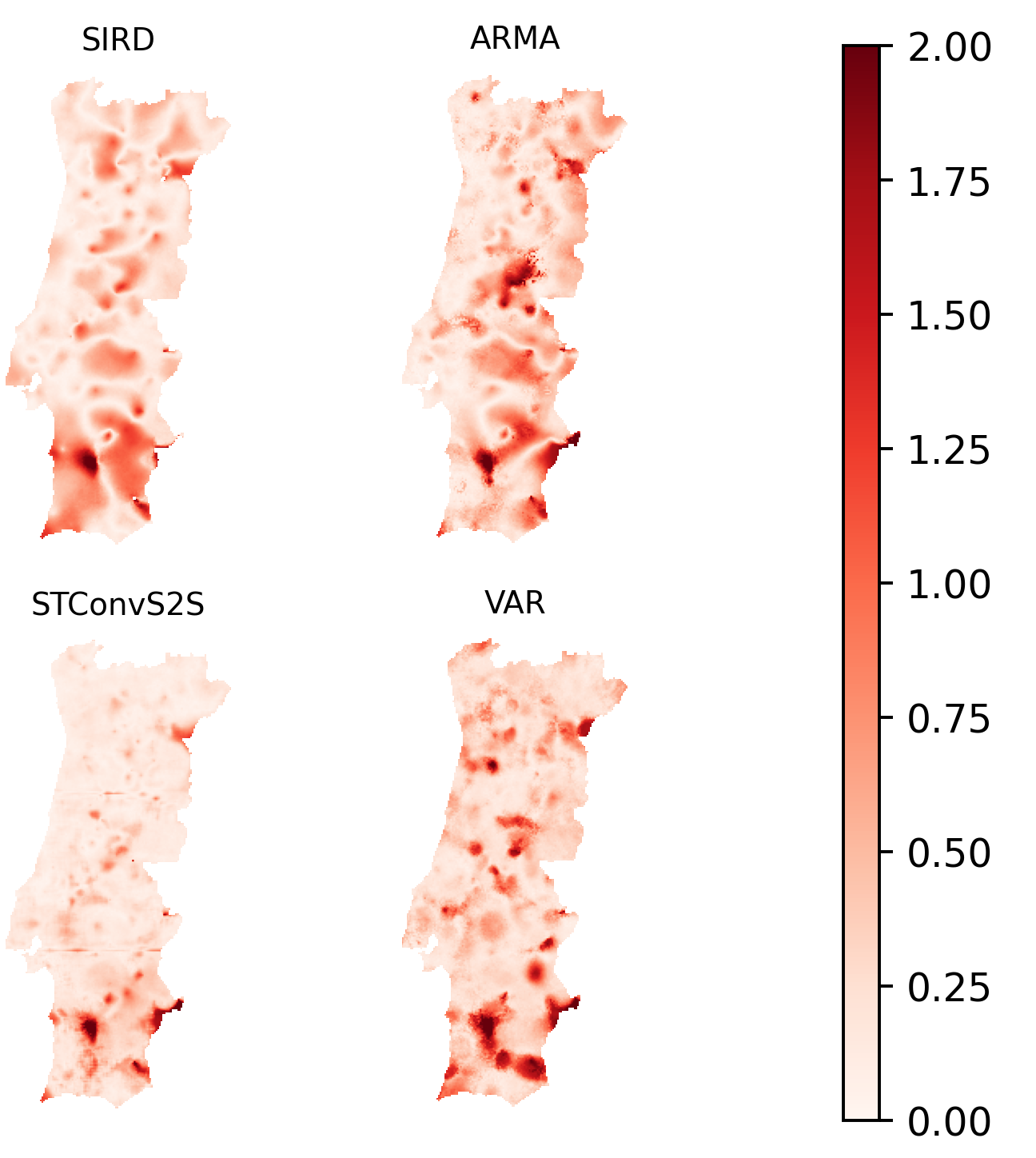}
\caption{Prediction error (sMAPE) for the VAR, ARMA, SIRD and STConvS2S models. \label{fig:heatmap2}}
\end{figure}

Figures \ref{fig:heatmap1} and \ref{fig:heatmap2} illustrates the quality of the predictions for week seventeenth of the predicted period, the week after Christmas  (December 28th 2020 - January 3rd 2021). This was a particularly relevant week, since it corresponded to a sharp inflection of the tendency and the start of the third wave. The predictions were made with the data available until December 21st, 2020, for the first day in the week. The window was then adjusted by one day for each successive day. The results clearly show the superior predictive ability of the STConvS2S model which, in this particular week, only exhibited significant error in a sparsely populated region in the south of Portugal, Alentejo, while the other models exhibited a much more significant error in different parts of the country. Other weeks exhibit similar behavior, although there is significant variation in the model errors over time, as shown in Figures \ref{fig:SMAPE} and \ref{fig:RMSE}.

\subsection{Analysis}

The results obtained in our tests have shown conclusively the superior predictive ability of the STConvS2S method, when compared with the other methods. We attribute this superiority to the ability of this model to take into consideration the incidence rate at neighboring cells, when predicting the evolution of that rate in a given cell. Although, conceptually, the VAR model could also use this information, it has no access to the position of each cell and, even though it beats the basic ARMA method, it does not reach the level of precision attained by the STConvS2S model. 

The SIRD model suffers from the same limitations as the ARMA model, in that it cannot use information from the neighboring municipalities to infer the dynamics of the pandemic. In principle, this model should have been the one better tuned to the reference data, since it uses exactly the same method to infer cell level incidence rates from municipality level rates as the approach used to obtain the reference data: the Block-DSS krigging algorithm. We attribute the lower precision of this method to two factors: the inherent limitations of the time-varying SIRD models to make predictions far in the future due to the changing pandemic dynamics; and the inability of the SIRD model to take into consideration the geographical relations between municipalities. We conjecture that a modified SIRD model that uses information from neighboring municipalities may improve significantly these results.

\section{Conclusions and future work \label{sec:conclusions}}

We computed an approximation to the spatio-temporal incidence rate of COVID-19 in mainland Portugal, using a methodology that uses the official municipality level information made available by the Portuguese Directorate-General for Healh (DGS) post-processed by direct block sequential simulation (block-DSS). This dataset is made available with the publication of this paper, and was used as a gold standard for the spatio-temporal evolution of the COVID-19 incidence rates during the first year, in this specific geography. 

We then used these data as a gold standard to test the behavior of four predictive models: an autoregressive-moving-average model, a vector autoregressive model, a model based on the combination of a SIRD compartmental model with block-DSS, and a model based on the STConvS2S methodology. We concluded that the STConvS2S neural network performed significantly better than the alternatives considered, and should therefore be considered the state-of-the-art for this problem. We invite researchers interested in applying spatio-temporal prediction methods to use these data, and to compare the predictions with the ones obtained by the four methods reported in this study.

All the data used in this paper, a significant set of additional results including the day-by-day previsions, maps and videos of incidence rate, as well as the code for the SIRD and the STConvS2S models is available at http://covid.vps.tecnico.ulisboa.pt/models. The baseline incidence rate data and the previsions of this model have been integrated in the interactive information site available at http://covid.vps.tecnico.ulisboa.pt.

\section{Acknowledgments}

The authors would like to thank André Peralta Santos and the Portuguese Directorate-General for Health, for making available the municipality level epidemiological data. This work was funded by projects INTAKE, SMOCK and MARÉ, financed by the Portuguese Science Foundation.

\bibliographystyle{IEEEtran}
\bibliography{references}

\begin{thebibliography}{10}
\providecommand{\url}[1]{#1}
\csname url@samestyle\endcsname
\providecommand{\newblock}{\relax}
\providecommand{\bibinfo}[2]{#2}
\providecommand{\BIBentrySTDinterwordspacing}{\spaceskip=0pt\relax}
\providecommand{\BIBentryALTinterwordstretchfactor}{4}
\providecommand{\BIBentryALTinterwordspacing}{\spaceskip=\fontdimen2\font plus
\BIBentryALTinterwordstretchfactor\fontdimen3\font minus
  \fontdimen4\font\relax}
\providecommand{\BIBforeignlanguage}[2]{{%
\expandafter\ifx\csname l@#1\endcsname\relax
\typeout{** WARNING: IEEEtran.bst: No hyphenation pattern has been}%
\typeout{** loaded for the language `#1'. Using the pattern for}%
\typeout{** the default language instead.}%
\else
\language=\csname l@#1\endcsname
\fi
#2}}
\providecommand{\BIBdecl}{\relax}
\BIBdecl

\bibitem{Azevedo20}
L.~Azevedo, M.~J. Pereira, M.~C. Ribeiro, and A.~Soares, ``Geostatistical
  {COVID}-19 infection risk maps for {P}ortugal,'' \emph{International Journal
  of Health Geographics}, vol.~19, no.~1, pp. 1--8, 2020.

\bibitem{scarpone2020multimethod}
C.~Scarpone, S.~T. Brinkmann, T.~Gro{\ss}e, D.~Sonnenwald, M.~Fuchs, and B.~B.
  Walker, ``A multimethod approach for county-scale geospatial analysis of
  emerging infectious diseases: a cross-sectional case study of {COVID}-19
  incidence in {G}ermany,'' \emph{International journal of health geographics},
  vol.~19, no.~1, pp. 1--17, 2020.

\bibitem{doblhammer2020social}
G.~Doblhammer, C.~Reinke, and D.~Kreft, ``Social disparities in the first wave
  of {COVID}-19 infections in {G}ermany: A county-scale explainable machine
  learning approach,'' \emph{medRxiv}, 2020.

\bibitem{miller2020mapping}
I.~F. Miller, A.~D. Becker, B.~T. Grenfell, and C.~J.~E. Metcalf, ``Mapping the
  burden of {COVID}-19 in the {U}nited {S}tates,'' \emph{medRxiv}, 2020.

\bibitem{gatrell1996spatial}
A.~C. Gatrell, T.~C. Bailey, P.~J. Diggle, and B.~S. Rowlingson, ``Spatial
  point pattern analysis and its application in geographical epidemiology,''
  \emph{Transactions of the Institute of British geographers}, pp. 256--274,
  1996.

\bibitem{de2020geospatial}
D.~De~Ridder, J.~Sandoval, N.~Vuilleumier, S.~Stringhini, H.~Spechbach,
  S.~Joost, L.~Kaiser, and I.~Guessous, ``Geospatial digital monitoring of
  {COVID}-19 cases at high spatiotemporal resolution,'' \emph{The Lancet
  Digital Health}, vol.~2, no.~8, pp. e393--e394, 2020.

\bibitem{kuo2018characterizing}
F.-Y. Kuo, T.-H. Wen, and C.~E. Sabel, ``Characterizing diffusion dynamics of
  disease clustering: A modified space--time {DBSCAN} ({MST-DBSCAN})
  algorithm,'' \emph{Annals of the American Association of Geographers}, vol.
  108, no.~4, pp. 1168--1186, 2018.

\bibitem{Oliveira2013stochastic}
A.~R. Oliveira, C.~Branquinho, M.~Pereira, and A.~Soares, ``Stochastic
  simulation model for the spatial characterization of lung cancer mortality
  risk and study of environmental factors,'' \emph{Mathematical Geosciences},
  vol.~45, no.~4, pp. 437--452, 2013.

\bibitem{waller2004applied}
L.~A. Waller and C.~A. Gotway, \emph{Applied spatial statistics for public
  health data}.\hskip 1em plus 0.5em minus 0.4em\relax John Wiley \& Sons,
  2004, vol. 368.

\bibitem{Goovaerts2005}
P.~Goovaerts, ``Geostatistical analysis of disease data: estimation of cancer
  mortality risk from empirical frequencies using {P}oisson kriging,''
  \emph{International Journal of Health Geographics}, vol.~4, no.~1, pp. 1--33,
  2005.

\bibitem{ross1916application}
R.~Ross, ``An application of the theory of probabilities to the study of a
  priori pathometry — {P}art i,'' \emph{Proceedings of the Royal Society of
  London. Series A, Containing papers of a mathematical and physical
  character}, vol.~92, no. 638, pp. 204--230, 1916.

\bibitem{kermack1927contribution}
W.~O. Kermack and A.~G. McKendrick, ``A contribution to the mathematical theory
  of epidemics,'' \emph{Proceedings of the royal society of london. Series A,
  Containing papers of a mathematical and physical character}, vol. 115, no.
  772, pp. 700--721, 1927.

\bibitem{bailey1975mathematical}
N.~T. Bailey \emph{et~al.}, \emph{The mathematical theory of infectious
  diseases and its applications}.\hskip 1em plus 0.5em minus 0.4em\relax
  Charles Griffin \& Company Ltd, 5a Crendon Street, High Wycombe, Bucks HP13
  6LE., 1975.

\bibitem{caccavo2020chinese}
D.~Caccavo, ``Chinese and {I}talian {COVID}-19 outbreaks can be correctly
  described by a modified {SIRD} model,'' \emph{medRxiv}, 2020.

\bibitem{calafiore2020time}
G.~C. Calafiore, C.~Novara, and C.~Possieri, ``A time-varying {SIRD} model for
  the {COVID}-19 contagion in {I}taly,'' \emph{Annual reviews in control},
  2020.

\bibitem{chatterjee2020studying}
S.~Chatterjee, A.~Sarkar, S.~Chatterjee, M.~Karmakar, and R.~Paul, ``Studying
  the progress of {COVID}-19 outbreak in {I}ndia using {SIRD} model,''
  \emph{Indian Journal of Physics}, pp. 1--17, 2020.

\bibitem{fernandez2020estimating}
J.~Fern{\'a}ndez-Villaverde and C.~I. Jones, ``Estimating and simulating a
  {SIRD} model of {COVID}-19 for many countries, states, and cities,'' National
  Bureau of Economic Research, Tech. Rep., 2020.

\bibitem{schmidhuber1997long}
J.~Schmidhuber and S.~Hochreiter, ``Long short-term memory,'' \emph{Neural
  Comput}, vol.~9, no.~8, pp. 1735--1780, 1997.

\bibitem{cho2014learning}
K.~Cho, B.~Van~Merri{\"e}nboer, C.~Gulcehre, D.~Bahdanau, F.~Bougares,
  H.~Schwenk, and Y.~Bengio, ``Learning phrase representations using {RNN}
  encoder-decoder for statistical machine translation,'' \emph{arXiv preprint
  arXiv:1406.1078}, 2014.

\bibitem{shi2015convolutional}
X.~Shi, Z.~Chen, H.~Wang, D.-Y. Yeung, W.-K. Wong, and W.-c. Woo,
  ``Convolutional {LSTM} network: A machine learning approach for precipitation
  nowcasting,'' \emph{arXiv preprint arXiv:1506.04214}, 2015.

\bibitem{hong2017psique}
S.~Hong, S.~Kim, M.~Joh, and S.-K. Song, ``Psique: Next sequence prediction of
  satellite images using a convolutional sequence-to-sequence network,''
  \emph{arXiv preprint arXiv:1711.10644}, 2017.

\bibitem{alleon2020plumenet}
A.~All{\'e}on, G.~Jauvion, B.~Quennehen, and D.~Lissmyr, ``Plumenet:
  Large-scale air quality forecasting using a convolutional lstm network,''
  \emph{arXiv preprint arXiv:2006.09204}, 2020.

\bibitem{wang2017predrnn}
Y.~Wang, M.~Long, J.~Wang, Z.~Gao, and P.~S. Yu, ``Predrnn: Recurrent neural
  networks for predictive learning using spatiotemporal {LSTM}s,'' in
  \emph{Proceedings of the 31st International Conference on Neural Information
  Processing Systems}, 2017, pp. 879--888.

\bibitem{castro2021stconvs2s}
R.~Castro, Y.~M. Souto, E.~Ogasawara, F.~Porto, and E.~Bezerra, ``{STConvS2S}:
  Spatiotemporal convolutional sequence to sequence network for weather
  forecasting,'' \emph{Neurocomputing}, vol. 426, pp. 285--298, 2021.

\bibitem{Liu09}
Y.~Liu and A.~G. Journel, ``A package for geostatistical integration of coarse
  and fine scale data,'' \emph{Computers \& Geosciences}, vol.~35, no.~3, pp.
  527--547, 2009.

\bibitem{whittle1963prediction}
P.~Whittle, \emph{Prediction and regulation by linear least-square
  methods}.\hskip 1em plus 0.5em minus 0.4em\relax Univ. of Minnesota Press,
  1963.

\bibitem{sims1980macroeconomics}
C.~A. Sims, ``Macroeconomics and reality,'' \emph{Econometrica: journal of the
  Econometric Society}, pp. 1--48, 1980.

\bibitem{liu2020evolving}
H.~Liu, A.~Brock, K.~Simonyan, and Q.~V. Le, ``Evolving
  normalization-activation layers,'' \emph{arXiv preprint arXiv:2004.02967},
  2020.

\bibitem{uselis2020localized}
A.~Uselis, M.~Luko{\v{s}}evi{\v{c}}ius, and L.~Stasytis, ``Localized
  convolutional neural networks for geospatial wind forecasting,''
  \emph{Energies}, vol.~13, no.~13, p. 3440, 2020.

\end{thebibliography}

\end{document}